\documentclass[runningheads]{llncs}

\usepackage{comment}
\usepackage{amsmath,amssymb} % define this before the line numbering.
\usepackage{color}

\usepackage{enumitem}
\usepackage{graphicx}
\usepackage{subfloat}
\usepackage{tabularx}
\usepackage{adjustbox}
\usepackage{amssymb}
\usepackage{multirow}
\usepackage[flushleft]{threeparttable}
\usepackage{subfigure}
\usepackage{tabularx}
\usepackage{rotating}
\usepackage{subfig}
\usepackage{blindtext}
\newcommand{\figref}[1]{Fig.~\ref{#1}}
\newcommand{\tabref}[1]{Table~\ref{#1}}

\newcommand{\sref}[1]{Sec.~\ref{#1}}

\begin{document}

\pagestyle{headings}
\mainmatter

\title{Gated Texture CNN \\for Efficient and Configurable Image Denoising} % Replace with your title

\titlerunning{Gated Texture CNN for Efficient and Configurable Image Denoising}

\author{Kaito Imai \and
Takamichi Miyata
}
\authorrunning{K. Imai and T. Miyata.}

\institute{ Chiba Institute of Technology, Chiba, Japan \\
\email{s16c3011fm@s.chibakoudai.jp,\\ takamichi.miyata@it-chiba.ac.jp}
}

\maketitle

\begin{abstract}

Convolutional neural network (CNN)-based image denoising methods typically estimate the noise component contained in a noisy input image and restore a clean image by subtracting the estimated noise from the input. However, previous denoising methods tend to remove high-frequency information (e.g., textures) from the input. It caused by intermediate feature maps of CNN contains texture information.
A straightforward approach to this problem is stacking numerous layers, which leads to a high computational cost.
To achieve high performance and computational efficiency, we propose a gated texture CNN (GTCNN), which is designed to carefully exclude the texture information from each intermediate feature map of the CNN by incorporating gating mechanisms.
Our GTCNN achieves state-of-the-art performance with 4.8 times fewer parameters than previous state-of-the-art methods.
Furthermore, the GTCNN allows us to interactively control the texture strength in the output image without any additional modules, training, or computational costs.

\end{abstract}

\section{Introduction}
\label{sec:1}
Image denoising, which is one of the fundamental tasks in computer vision, aims to recover an original image from a noisy observation and plays a vital role in a preprocessing step in high-level vision tasks (e.g., image recognition \cite{meetsHigh,connectHigh} and image retrieval \cite{Intro2}). Due to the limitations of computational resources, these real-world applications require an accurate and computationally efficient image denoising algorithm.

Recently, deep convolutional neural networks (CNNs) have been very successful at image denoising. Zhang et al. proposed a denoising CNN (DnCNN) \cite{DNCNN}, which is a stack of units consisting of a CNN, batch normalization \cite{BN}, and a rectified linear unit (ReLU) \cite{ReLu}, called a CBR unit, with global residual learning. The DnCNN outperforms the previous non-learning-based state-of-the-art methods \cite{NLBay,BM3D,WNNM} and is well suited for parallel computation on GPUs.
The great success of DnCNN has inspired much work on CNN-based denoising \cite{FFDNet,MemNet,RED,IRCN,Remez2018,N3Net,UCNN,NLRN,RidNet,RNAN,MWCNN,SGN,FOCNet}. These methods employ many layers and a massive number of parameters to achieve a high denoising performance compared with the DnCNN method.

\begin{figure}[t]
  \centering
  \includegraphics[width=\linewidth]{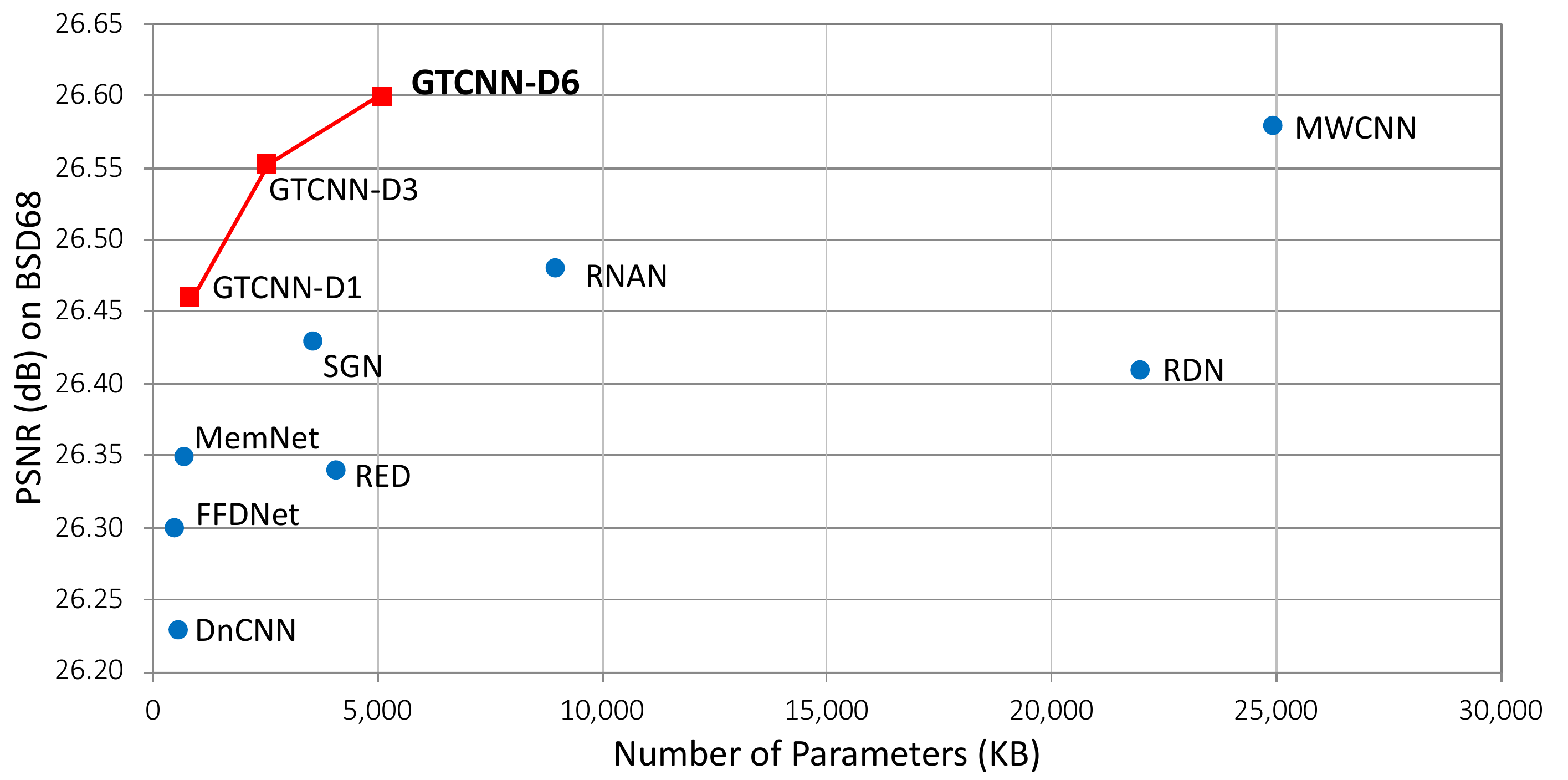}
    \caption{\textbf{Model size vs the denoising accuracy (PSNR)} with a noise level of $\sigma$ = 50. Our GTCNN significantly outperforms the other methods. In particular, GTCNN-D6 achieves new state-of-the-art denoising performance but is 4.8 times smaller than MWCNN \cite{MWCNN}  (see \tabref{table:sota} for more details)}

  \label{fig:fig1}

\end{figure}

The common problem of these methods is that they tend to remove the high-frequency information (e.g., textures, edge), which has similar properties to the noise, in their denoising process. General CNN-based denoising methods learn to estimate added noise from a noisy observation, which means that the intermediate feature maps (throughout this paper, we simply denote these maps as features) in those CNNs should contain only the noise information. However, these features unintentionally contain texture information, which leads to the removal of textures from noisy observations and worsens the denoising accuracy.

A straightforward approach to this problem is stacking more layers to eliminate the texture information carefully from the features. However, this simple approach suffers from high computational cost and time consumption and makes those denoisers infeasible to real-world applications. For example, one of the state-of-the-art denoising methods, a multilevel wavelet CNN (MWCNN) \cite{MWCNN}, has 43 times the number of parameters that the DnCNN has \cite{DNCNN} (see \figref{fig:fig1}). Then, the question arises: do we need such extremely deep and complex networks to remove the texture information from the features?

On the other hand, several image inpainting methods \cite{Partial,FF,chang2019free} incorporate a gating mechanism to mask the intermediate features based on the user's input. The gating mechanism allows us to choose which part of the features should be propagated to the next layer. With gating, these methods significantly improve the inpainting results with the user's guide.

Inspired by these results, we propose a novel denoising architecture referred to as a gated texture CNN (GTCNN) that can estimate the texture information contained in the features and remove them by a gating mechanism. We incorporate the gating texture layer (GTL) into the noise stream, as illustrated in \figref{fig:overall}. We employ a general denoising architecture for our noise stream to estimate the noise component from the noisy observations.
To estimate the texture, we have to capture the information from nearly the entire input image. To capture such information, we use a U-Net-like network \cite{UNet}, which has a wider receptive field than the noise stream, for the proposed GTL.

Our extensive evaluation shows that the proposed GTCNN enables us to scale the model with a good trade-off between accuracy and speed and that it is suitable for resource-limited devices. \figref{fig:fig1} shows the model size and the performance comparison on the 68-image Berkley Segmentation Dataset (BSD68) \cite{BSD68} with a high level of noise. Our GTCNN-D6 (with a heavyweight configuration) outperforms the state-of-the-art models with 4.8 times fewer parameters than the previous best model \cite{MWCNN}. Notably, GTCNN-D1 (with a lightweight configuration) outperforms previously proposed efficient denoising methods \cite{DNCNN,FFDNet,SGN,RED,MemNet} in terms of the trade-off between the denoising performance, speed, and parameter efficiency.

The experimental results show not only the superior performance but also that our GTCNN model allows users to control the texture strength of the denoising results by tuning a single parameter in the GTL (see \sref{sec:comp_dni}). Interactive adjustment of the restoration strength is a crucial feature in real-world image processing applications, such as photo editing. Moreover, these results show that our GTCNN can separate noise and textures at the intermediate level.
To the best of our knowledge, this is the first study to show that the gate mechanism can separate textures from features. Moreover, this is the first study to determine the architecture that allows users to control the texture strength in image denoising without sacrificing the performance and computational efficiency.

\smallskip\noindent\textbf{Contributions.} Our main contributions are threefold.
\begin{enumerate}[topsep=0pt,itemsep=0pt]
\item We show that a simple gating mechanism that removes textures from the intermediate feature maps is effective for image denoising. Our GTCNN achieves state-of-the-art performance in terms of the efficiency and denoising performance.
\item We show that the GTL can effectively separate textures from CNN feature maps.
\item Our GTCNN allows users to interactively and continuously control the texture strength in the denoising results, which suggests that our GTL can remove texture from noise features.
\end{enumerate}

\section{Related Work}

\subsection{CNN-based Image Denoising Methods}
Zhang et al. \cite{DNCNN} revealed that the combination of batch normalization and a global residual skip connection, which estimates the residuals between the noisy input and the corresponding clean image, plays a key role in CNN-based image denoising methods. Their proposed architecture, DnCNN, clearly outperforms previous, non-learning-based denoising methods such as the block-matching and 3D filtering (BM3D) \cite{BM3D}, non-local Bayes \cite{NLBay}, and weighted nuclear norm minimization (WNNM) \cite{WNNM} algorithms. The DnCNN algorithm is widely accepted as the baseline for CNN-based denoising methods.

To achieve a high denoising performance, more complex networks \cite{RNAN,NLRN,MWCNN,RDN,N3Net} have been proposed.
A residual dense network (RDN) \cite{RDN} incorporates densely connected convolutional networks \cite{DenceNet} for image restoration. Some methods \cite{RNAN,NLRN} use a nonlocal (NL) module \cite{NL} designed in the neural network. Zhang et al. proposed the residual NL attention network (RNAN) \cite{RNAN}, which incorporates NL attention blocks \cite{NL} and residual attention learning \cite{RArecog} to capture global information.
Liu et al.  proposed the MWCNN algorithm \cite{MWCNN}, which first employs multilevel wavelet pooling to avoid information loss.

These recent CNN-based methods are powerful but lead to high computational costs. For example, MWCNN has 43 times the parameters and is 3.5 times slower than DnCNN. Gu et al. proposed the self-guided network (SGN) \cite{SGN} for fast image denoising and achieved a better trade-off between the denoising accuracy and runtime by a pixel shuffle operation \cite{pixelshuffle}. However, the SGN needs a large number of parameters to achieve a good trade-off.

\subsection{Gating Mechanism}

Gating mechanisms were originally proposed and applied to recurrent neural networks in natural language processing \cite{LSTM}. Dauphin et al. incorporated gating mechanisms into CNNs and replaced recurrent neural networks with gated convolutions for language modeling \cite{GateCNN}.

Gating mechanisms have also been applied to many computer vision problems \cite{SENet,FF,Gateseg,PixelCNN,GSCNN,chang2019free,Veit2018} for controlling which information should be passed to the next layer.
For example, Yu et al. proposed the use of convolutions with a soft-gating mechanism for user-guided image inpainting to control feature flow in networks \cite{FF}. Takikawa et al. incorporated a gating mechanism with multitask learning to extract shape information for more accurate image segmentation \cite{GSCNN}.
In this paper, we propose using a gating mechanism to remove the texture information from the features for image denoising.

\subsection{Interactive Modulation for Denoising Results }

CNN-based methods lack the flexibility to control their denoising results. In real-world applications, flexibility is equally essential as denoising accuracy and computational efficiency.
Recently, interactive control of denoising output has attracted increasing attention in the image restoration field \cite{AdaFM,DNI,Dynamic-Net,CFSNet}.

Several recent works \cite{AdaFM,DNI,Dynamic-Net,CFSNet} for interactive modulation proposed incorporating an additional network into the denoising CNN, such as DnCNN, to allow control over the output results.
For example, Wang et al. proposed deep network interpolation (DNI), which allows the interpolation of the denoising strength between several networks trained by images with different denoising strengths \cite{DNI}.
However, this network interpolation strategy requires at least several networks for interpolation, which leads to a heavy computational cost that is unacceptable for resource-limited devices.
Our GTCNN allows us to continuously control the strength of the texture information in the denoising results without any additional training, modification modules, or postprocessing steps.

% -----------------------------------------------------------------------
% ------------------------------Gated Texture CNN-----------------------------------------
% -----------------------------------------------------------------------

\section{Gated Texture CNN}
\begin{figure}[t]
  \centering
  \includegraphics[width=\linewidth]{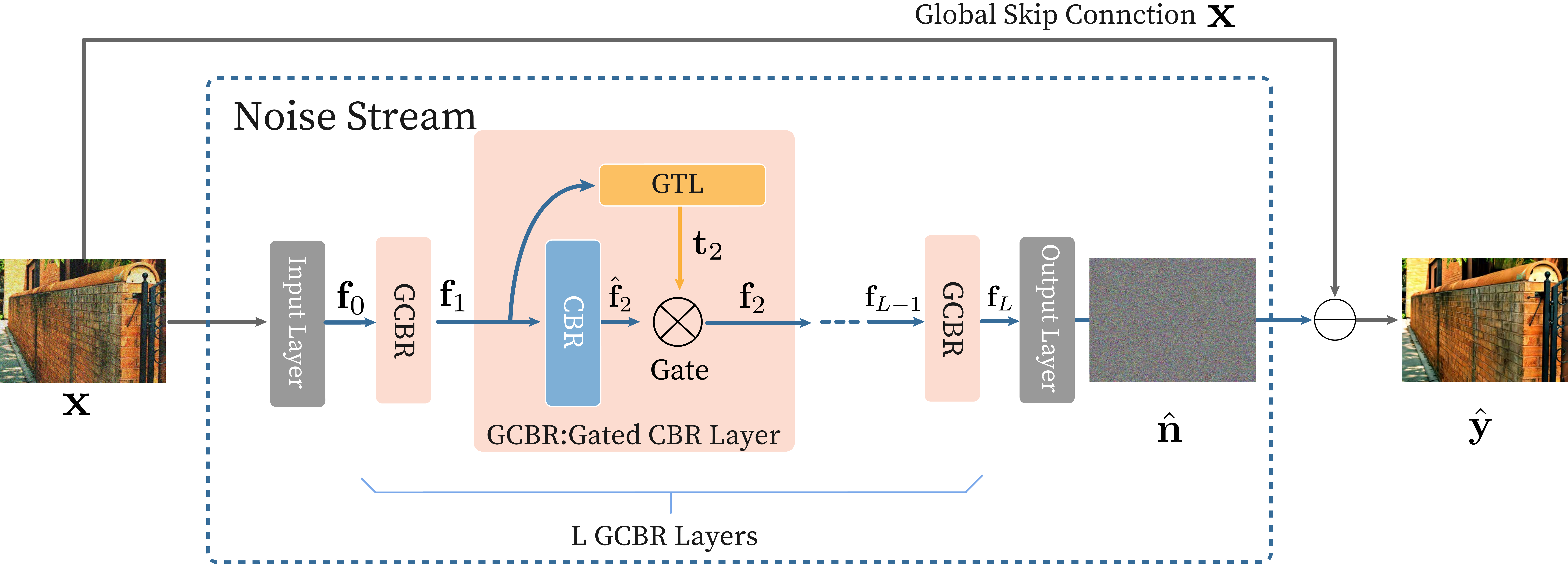}
  \caption{\textbf{Overview of the GTCNN architecture.} To remove the texture information from the intermediate features, we incorporate our gated texture layer (GTL) into the CBR units with a gating mechanism }

  \label{fig:overall}
\end{figure}
\def\e{{\rm \mathbf e}}
\def\n{{\mathbf n}}
\def\x{{\mathbf x}}
\def\y{{\mathbf y}}
\def\f{{\mathbf f}}
\def\e{{\mathbf e}}
\def\t{{\mathbf t}}
\def\inwhc{{\in \mathbb{R}^{H \times W \times C}}}
\def\inwhci{{\in \mathbb{R}^{H \times W \times C_I}}}
\def\whc{{\mathbb{R}^{H \times W \times C}}}
\def\inwh{{\in \mathbb{R}^{H \times W}}}

First, we introduce a standard noise stream based on the DnCNN, which estimates the added noise of the input image (\sref{sec:31}). Then, we describe a novel GTL (\sref{sec:32}) and how to modify the strength of the texture component in the denoising results (\sref{sec:33}).

\subsection{Noise Stream}
\label{sec:31}
Let $H$, $W$, and $C_I$ be height, width, and number of the channels of the image. A noisy observation image $\x \inwhci$ can be modeled as $\x=\y + \n$, where $\y$ refers to the unknown clean image, $\n$ is additive white Gaussian noise.

The purpose of the noise stream is to obtain a $ \hat \n \inwhci$ as the estimate of $\n$ from the input and then obtain a latent clean image $ \hat \y \inwhci$ by subtracting $ \hat \n$ from $\x$.
The noise stream consists of an input/output layer and $L$ intermediate layers,  as illustrated in \figref{fig:overall}. The input layer converts the noisy input image into the first intermediate features $\f_0 \inwhc$, where $C$ is the number of channels in the feature vectors.

The intermediate layers play crucial roles in estimating noise, where each layer takes the previous features $\f_{l-1} \inwhc$ and $(l= 0, \dots, L)$ as its input and outputs the new features $\f_{l}$. The output layer produces the estimated noise image $ \hat \n $. The final output is obtained by the global residual connection as $ \hat \y = \x - \hat \n$.

Ideally, the final features $\f_{L}$ should contain only the noise information. However, as we mentioned in \sref{sec:1}, the final features of the previous methods using the CBR units still contain some texture information, which might be the cause a drop in performance.
We replace this intermediate CBR layer with our gated CBR (GCBR) layer, which employs our gated texture layer (GTL) to remove the texture information from the features.

\subsection{GCBR Layer}
\label{sec:32}

Given an intermediate feature map $\f_{l-1}$ as input, the GCBR layer infers the noise features $\f_{l} \inwhc $ and the texture features $\t_{l} \inwhc$ in parallel, as illustrated in \figref{fig:overall}.
Then, the GCBR layer removes the texture information $\t_{l}$ from the noise features $\hat \f_{l}$ by the gating mechanisms. We define the process of the GCBR layer as $\phi$: $ \whc \xrightarrow{}  \whc$.

The overall GCBR layer process can be summarized as:
\begin{equation}\label{eq:GCBR}
\begin{split}
    \phi (\f_{l-1})  &= \hat \f_{l} \otimes \t_{l},
\end{split}
\end{equation}
where  $\otimes$ denotes Hadamard multiplication. $\hat \f_{l}$ is given by

\begin{equation}\label{eq:CBR}
\begin{split}
    \hat \f_{l}  &= \theta(\f_{l-1}),
\end{split}
\end{equation}
where $\theta: \whc \xrightarrow{}  \whc$ is a map corresponding to a CBR layer.

\subsubsection{Gated Texture Layer.}

Our GTL is a crucial component of our GTCNN that estimates texture information from the input features to remove the texture information.
We assume that the network, which estimates the texture information from the noise, should have a large receptive field to capture the global context of the scene.
We empirically reveal that the large receptive field significantly improves the denoising performance (see \sref{sec:rec_abl}).
From this assumption, we design our GTL based on U-Net \cite{UNet}, as illustrated in \figref{fig:GTL}.
U-Net is an encoder-decoder network with a skip connection between the corresponding layers of the encoder and decoder of the same stage. Note that the layers of each stage are composed of two CBR layers; we call these layers double CBR (DCBR) layers.

The general U-Net decreases the size of the features by a max-pooling operation and increases the number of channels at each stage.
The pooling operation enlarges the receptive field and allows us to capture the context. However, increasing the number of channels does not guarantee an improvement in the denoising performance, although it leads to a high computational cost.

We modify our GTL so that it does not increase the number of channels. This modification makes our GTL more computationally efficient than the original U-Net \cite{UNet}, as it has five times fewer parameters then the original U-Net.
Furthermore, we found that increasing the depth $L$ of the noise stream above a certain level would make the training procedure numerically unstable. We resolve this issue by placing a $ 1 \times 1 $ convolution layer after the last DCBR layer of the decoder. This layer stabilizes the training procedure greatly when using the deep noise stream.

The GTL takes the previous features $ \f_{l-1}$ as input and estimates the texture information.
The final estimate of the texture information is obtained by:
\begin{equation}
\t_{l} =  \delta( \gamma(\f_{l-1})),
\label{eq:GTL}
\end{equation}
where $\gamma $ is the process of the GTL ($\gamma $: $\whc \xrightarrow{} \whc$) and $\delta$ is the channelwise softmax function.
The sigmoid function was commonly employed in previous works that included gating mechanisms \cite{SENet,FF,Gateseg,PixelCNN,GSCNN,Veit2018}.

We found that the softmax function achieve a better result than the sigmoid function in removing the texture information. We will show the experimental results of the ablation study in \sref{sec:ablation} to illustrate our findings.

\subsection{Modification}
\begin{figure}[t]
  \centering
  \includegraphics[width=\linewidth]{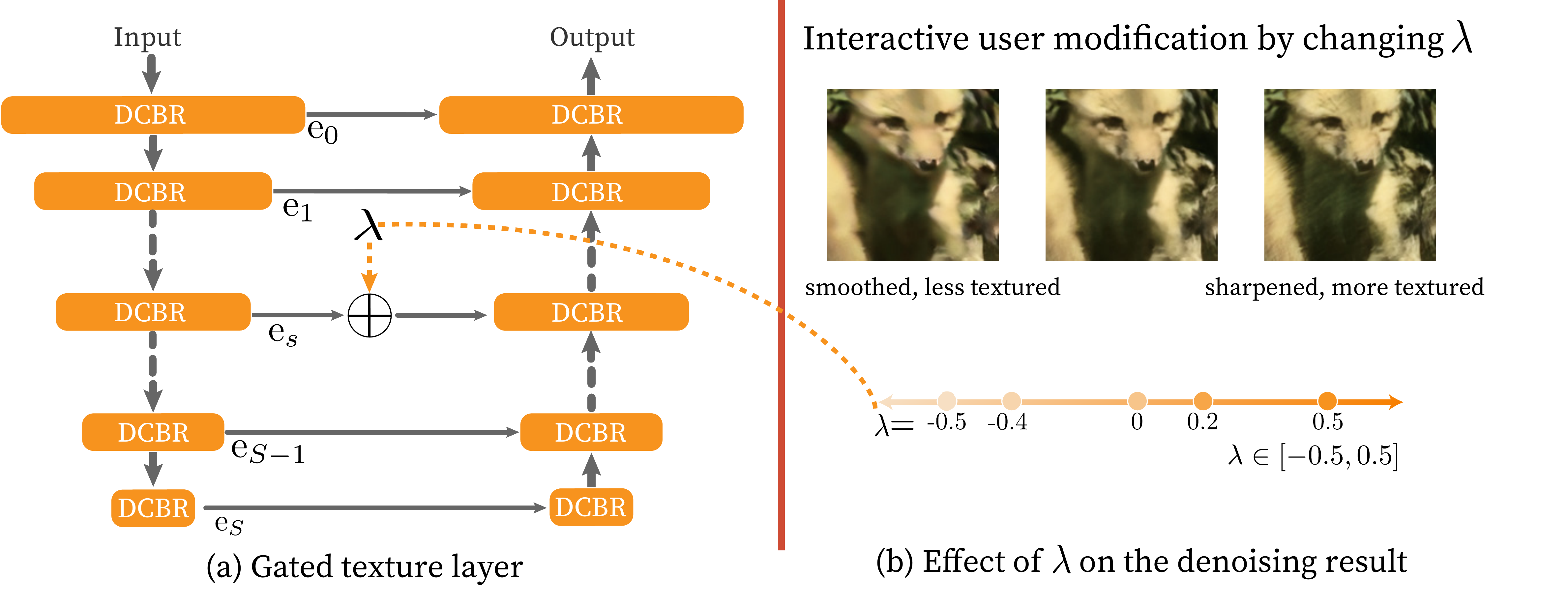}
  \caption{\textbf{GTL architecture and how to modulate the texture strength.} (a) To capture the global context of the scene, we design our gated texture layer (GTL) based on U-Net \cite{UNet}. (b) Our GTCNN allows users to interactively and continuously control the texture gating strength by just shifting an output of the skip connection in GTL}

  \label{fig:GTL}
\end{figure}
\label{sec:33}
For a given input, most CNN-based denoising methods can generate only a single fixed output. In other words, these methods have no flexibility in controlling the denoising results. However, in real-world applications, flexibility is as essential as the denoising accuracy and computational efficiency.
Several recent works \cite{AdaFM,DNI,CFSNet} incorporated an additional network to the denoising CNN, such as DnCNN, to enable control over the output results. However, there was a trade-off between the flexibility and computational efficiency.

Since our GTCNN is designed to separate the texture information from the noise stream, we can expect that modulating the GTL will allow us to control the texture  gating strength in the denoising results.
We show that the network output could change continuously by just modulating the output of the skip connection in the GTL.

We define $\e_s \inwhc$ as the output of the skip connection of the $s$th stage. To control the texture gating strength, we just need to perform an elementwise shift $\e_s$ by a shifting parameter of $\lambda \in [-0.5,0.5]$, as illustrated in \figref{fig:GTL}-b.

When we modulate $\lambda$ gradually from -0.5 to 0.5 on the inference time, the network output will change continuously, as shown in \figref{fig:mod_comp_dni}. A detailed analysis can be found in \sref{sec:mod_abl}.
Note that unlike previous modification methods \cite{AdaFM,DNI,CFSNet}, this modulation does not need any additional training, modification modules, or postprocessing steps. Furthermore, we do not sacrifice the computational efficiency or the denoising accuracy.

% -----------------------------------------------------------------------
% ------------------------------Experiments studies-----------------------------------------
% -----------------------------------------------------------------------
\section{Experiments}

To thoroughly evaluate the effectiveness of our GTCNN, we first compare our method with the current state-of-the-art denoising methods \cite{MWCNN,RNAN,RDN,SGN,MemNet,RED}, which require high computational resources. Then, we also compare our GTCNN with several computationally efficient CNN-based methods \cite{DNCNN,FFDNet,SGN,MemNet,RED} in terms of the denoising performance in grayscale and color images. Furthermore, we show the configurability of the GTCNN by comparing it with a previous feature modification method \cite{AdaFM}. We also performed extensive ablation studies to show the effectiveness of our design choice.

For a fair comparison, we used the authors' official implementations in PyTorch \cite{Pytorch} and the pre-trained models to reproduce the results of the comparison methods. If the authors not provide the pre-trained models, we trained those methods by ourselves. If necessary, we also used some unofficial PyTorch \cite{Pytorch} implementations that are linked from the original authors' GitHub repositories. For all the methods, we use the default settings provided by the respective authors. We confirmed that the maximum absolute difference between our reproduced results and the results provided in the original papers is only 0.01 dB.

\subsection{Experimental Setup}
We use the experimental settings from previous works \cite{MWCNN,RNAN,SGN}.

\subsubsection{Network Settings.}

We denote a GTCNN with $L$ intermediate (GCBR) layers as GTCNN-D$L$. For example, GTCNN-D1 means a GTCNN with only one GCBR layer. In the evaluation, we used GTCNN-D1/D3/D6. Note that GTCNN-D6 employs a $1 \times 1$ convolutional layer after the final DCBR layer of the GTL for stable training, as we mentioned in \sref{sec:32}. All the GTCNN models use the same number of stages ($S=4$) for the GTL.

\subsubsection{Training Settings.}
We use the 800 standard 2K resolution
training images of the DIVerse 2K (DIV2K) dataset \cite{DIV2k}. We extracted $192 \times 192$ patches from the training images as a training set. The size of the stride is 192. The training set consisted of 55,500 patches. Unlike previous works \cite{MWCNN,RNAN,SGN}, we did not perform any data augmentation on the training set (e.g., random flipping).

Our model is trained by the Adam optimizer
\cite{Adam} with a learning rate of $0.001$, along with the cosine annealing technique proposed by \cite{Cos} for adjusting the learning rate. We conducted all the experiments in PyTorch \cite{Pytorch}.

\subsubsection{Evaluation Datasets and Metrics.}
To compare with the previous denoising algorithms, we use the commonly used BSD68 \cite{BSD68} dataset, which contains 68 images, and the Urban100 dataset \cite{Urban100}. We also use Set12 \cite{DNCNN} for the grayscale image denoising experiments.

The denoising results are evaluated with the PSNR. To evaluate the computational efficiency, we provide comparisons on the number of parameters. We also give the runtime for processing a 320 $ \times $ 480 image for reference. Note that the runtime is dependent on the computing environment and implementation. As we mentioned before, we used the PyTorch \cite{Pytorch} implementations published or introduced by the respective authors for all the comparison methods.

% -----------------------------------------------------------------------
% ------------------------------Compared with the state-of-the-art-----------------------------------------
% -----------------------------------------------------------------------

\begin{table}[t]
\setlength{\tabcolsep}{1mm} 
\caption{\textbf{GTCNN performance results}. The denoising methods with similar performance on Set12 are grouped together for effective comparisons. Our scaled GTCNN consistently reduced the number parameters and the runtime. Note that, even in a heavyweight configuration such as GTCNN-D6, the number of parameters and runtime do not increase rapidly} 

\begin{center}
\begin{tabular}{ l|c|c|c|c|c }
\hline

Method& Set12 & BSD68 & Urban100 & Number of Parameters & Time [ms]  \\
\hline
\hline
MemNet \cite{MemNet}      & 27.39 &  26.35 & 26.67   &  \textbf{685}k & 169.1  \\                       
                                    
RED \cite{RED}      & 27.35 & 26.34 & 26.46   &  4,100k & 41.3 \\                                              

SGN \cite{SGN}     & 27.53 & 26.43 & 26.96 & 3,577k & \textbf{7.5}
 \\
GTCNN-D1 (ours)       & \textbf{27.56} & \textbf{26.46} & \textbf{26.97} & 851k & 12.4 \\
\hline
\hline
RDN \cite{RDN}      & 27.60 & 26.41 & 27.40 & 21,973k & 716.8 \\

RNAN \cite{RNAN}      & 27.67 & 26.47 & \textbf{27.65} & 8,957k & 1,434.9 \\

GTCNN-D3 (ours)       & \textbf{27.76}& \textbf{26.55} & 27.50 & \textbf{2,552}k & \textbf{35.3}\\
\hline
\hline
MWCNN  \cite{MWCNN}     & 27.79 & 26.58 &  27.53 & 24,927k & 78.2 \\                                              
GTCNN-D6 (ours)      & \textbf{27.83} & \textbf{26.60}  & \textbf{27.72} & \textbf{5,128}k & \textbf{72.7} \\

\hline
\end{tabular}
\end{center}

\label{table:sota}
\end{table}

\subsection{Comparison with the State-of-the-Art Methods}
\label{sec:comp_sota}
We performed a comparison with the current state-of-the-art methods, namely, deep residual
encoder-decoder (RED) \cite{RED}, MemNet \cite{MemNet}, RNAN \cite{RNAN}, RDN \cite{RDN}, SGN \cite{SGN}, and MWCNN \cite{MWCNN}, with grayscale images under severe noise ($\sigma$=50).
\tabref{table:sota} illustrates the parameters and the denoising accuracy of all GTCNN models. Our GTCNN models generally use an order of magnitude fewer parameters than other CNN-based methods and achieve a similar accuracy. In particular, our GTCNN-D6 achieves the state-of-the-art result with 5,128k parameters, which is 4.8 times fewer than the previous best MWCNN model \cite{MWCNN}. Notably, our GTCNN models are not only small-sized but also fast. For example, our GTCNN-D3 achieves a comparable accuracy with RNAN \cite{RNAN}, but the GTCNN-D3 model is 40.6 times faster and has 2.8 times fewer parameters than the RNAN model. Note that the RNAN algorithm \cite{RNAN} uses NL operations \cite{NL}, which leads to high computational memory requirements and high time consumption.

\begin{figure}[t]
  \centering
  \includegraphics[width=\linewidth]{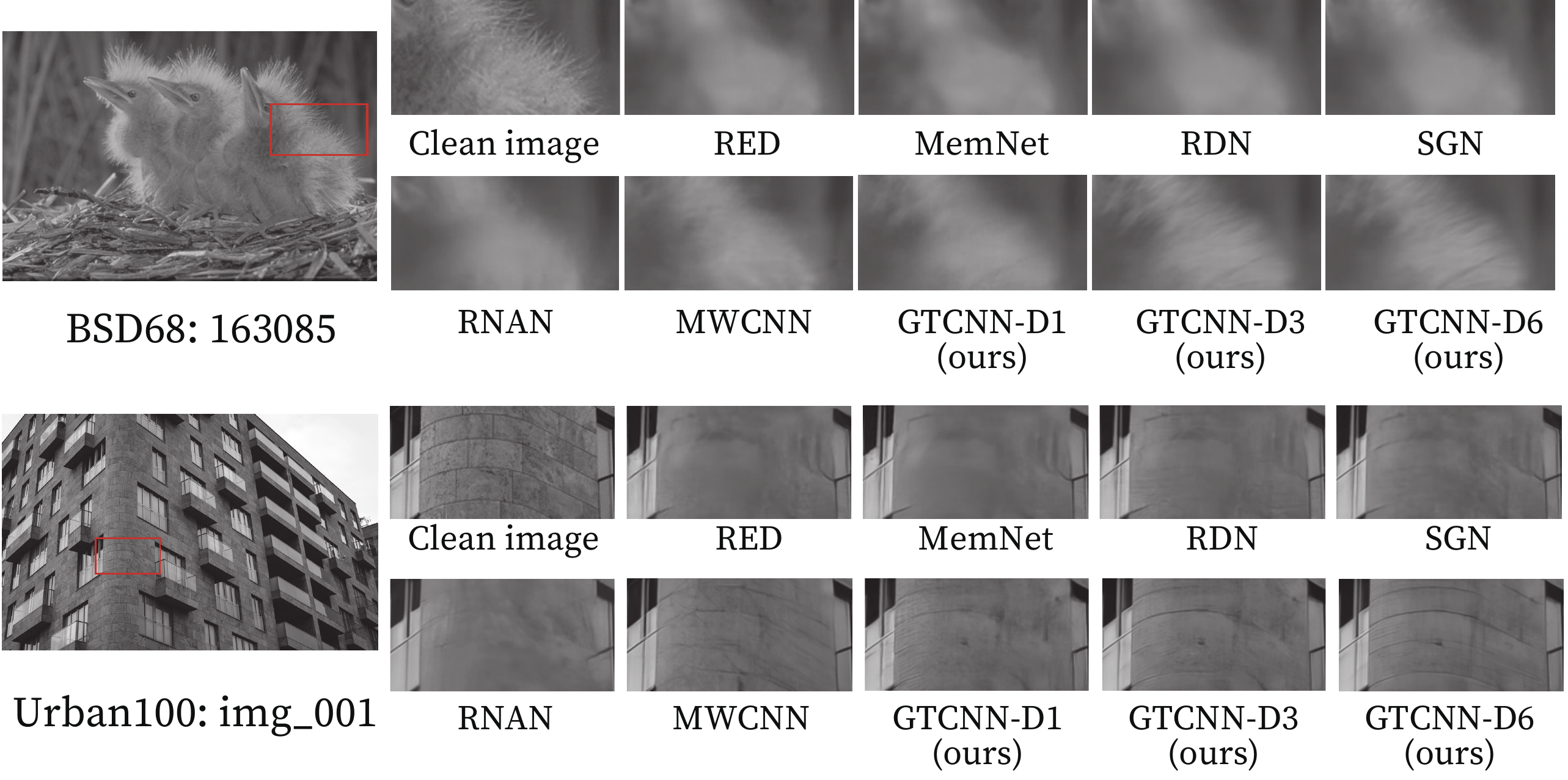}
    \caption{\textbf{Grayscale} image denoising results with a noise level of $\sigma$ = 50}

  \label{fig:Graycomp}

\end{figure}

The visual comparisons are shown in \figref{fig:Graycomp}.
The other methods in the comparison remove details along with the noise, which results in oversmoothing artifacts. Our texture gating strategy can restore sharp textures from noisy input without artifacts.

%  ---------------------------figure comp---------------------------------------

%  ---------------------------figure comp---------------------------------------

% -----------------------------------------------------------------------
% ------------------------------Compare-----------------------------------------
% -----------------------------------------------------------------------
\subsection{Comparison with Computationally Efficient Methods}
\label{sec:comp_efficent}
We compared our GTCNN model with previous methods that have relatively smaller model sizes, such as the DnCNN \cite{DNCNN}, fast and flexible denoising CNN (FFDNet) \cite{FFDNet}, RED \cite{RED}, and SGN \cite{SGN} models. For these experiments, we use additive white Gaussian noise with standard deviations $\sigma$ of 30, 50, and 70.
We show that the denoising results for the grayscale and color images in \tabref{tab:resultspsnrgray} and \tabref{tab:results_psnr_denoise_rgb}, respectively.
We also provide the number of parameters and runtime in \tabref{table:speedparam}.

These results show that our GTCNN-D1 model outperforms all other methods in terms of the denoising accuracy. The GTCNN-D1 model also archives a better trade-off between the model size, inference time, and denoising accuracy than the other methods. The GTCNN-D1 model has 4.2 times fewer parameters but also a higher denoising accuracy than the SGN model.

%% Tabel 6, EVAL : Reslut of GRAY

\begin{table}[t]
\caption{Quantitative results from \textbf{grayscale} image denoising. The best results are \textbf{in bold}}
\begin{center}
\label{tab:resultspsnrgray}
\begin{tabular}{|l|c|c|c|c|c|c|c|c|c|c|c|c|c|}
\hline
\multirow{2}{*}{Method} &  \multicolumn{3}{c|}{Set12} &  \multicolumn{3}{c|}{BSD68} &  \multicolumn{3}{c|}{Urban100}   
\\
\cline{2-10}
 & 30 & 50 & 70  & 30 & 50 & 70  & 30 & 50 & 70 
\\
\hline
\hline

\hline
DnCNN \cite{DNCNN} 
     & 29.53
      & 27.19
       & 25.52

     & 28.36
      & 26.23
       & 24.90
 
         & 28.88
          & 26.27
           & 24.36
            
\\
\hline
FFDNet \cite{FFDNet}

 & 29.61
  & 27.32
   & 25.81

     & 28.39
      & 26.30
       & 25.04

         & 29.03
          & 26.51
           & 24.86
\\
\hline
MemNet \cite{MemNet} 

 & 29.63
  & 27.39
   & 25.90

     & 28.43
      & 26.35
       & 25.09

         & 29.10
          & 26.67
           & 25.01
\\
\hline
RED \cite{RED}

 & 29.70
  & 27.35
   & 25.80

     & 28.50
      & 26.34
       & 25.10

         & 29.18
          & 26.46
           & 24.82

\\
\hline

SGN \cite{SGN}

 & 29.77
  & 27.53
   & 25.90

     & 28.50
      & 26.43
       & 25.17

         & 29.41
          & 26.96
           & 25.29

\\
\hline
GTCNN-D1 (ours)
 & \textbf{29.80}
  & \textbf{27.56}
  & \textbf{26.08}

     & \textbf{28.53}
      & \textbf{26.46}
      & \textbf{25.21}

         & \textbf{29.43}
          & \textbf{26.97}
          & \textbf{25.36}

\\
\hline         
\end{tabular}
\end{center}

\end{table}

% %% Tabel 6, Comp 3 : color Denoising-----------

\begin{table}[t]
\caption{Quantitative results from \textbf{color} image denoising. The best results are \textbf{in bold}}
\begin{center}
\begin{adjustbox}{max width=\textwidth}

\begin{tabular}{|l|c|c|c|c|c|c|c|c|c|c|}
\hline
\multirow{2}{*}{Method} &    \multicolumn{3}{c|}{BSD68} &  \multicolumn{3}{c|}{Urban100}   
\\
\cline{2-7}
& 30 & 50 & 70  & 30 & 50 & 70 
\\
\hline
\hline

\hline
DnCNN \cite{DNCNN}

     & 30.40
      & 28.01
       & 26.56

         & 30.28
          & 28.16
           & 26.17
            
\\
\hline
FFDNet \cite{FFDNet}

     & 30.31
      & 27.96
       & 26.53

         & 30.53
          & 28.05
           & 26.39
\\
\hline
MemNet \cite{MemNet}

     & 28.39
      & 26.33
       & 25.08

         & 28.93
          & 26.53
           & 24.93
\\
\hline
RED \cite{RED}

     & 28.46
      & 26.35
       & 25.09

         & 29.02
          & 26.40
           & 24.74  

\\
\hline

SGN \cite{SGN}

         & 30.45
          & 28.18
           & 26.79

     & 30.75
      & 28.36
       & 26.85
\\
\hline
GTCNN-D1 (ours)

     & \textbf{30.51}
      & \textbf{28.24}
      & \textbf{26.83}

         & \textbf{30.90}
          & \textbf{28.53}
          & \textbf{26.90}
           
\\
\hline

\end{tabular}
\end{adjustbox}
\end{center}

\label{tab:results_psnr_denoise_rgb}
\end{table}

%% Tabel 6, Comp 3 : Speed and param
\begin{table}[!t]
\caption{\textbf{Comparison of the computational costs}. The number of parameters and runtime [ms] for processing a 480 $\times$ 320 image are shown for the different methods. All the methods were implemented in PyTorch \cite{Pytorch}, and the runtime was evaluated on system equipped with an NVIDIA GTX 1080 Ti GPU} 

\begin{center}
\begin{adjustbox}{max width=\textwidth}
\begin{tabular}{ l|c|c|c|c|c|c  }
\hline
Method &  DnCNN \cite{DNCNN}& FFDNet \cite{FFDNet} & MemNet \cite{MemNet}& RED \cite{RED}& SGN \cite{SGN} & GTCNN-D1 \\
\hline
\hline
Number of Parameters       & 555k & 485k  & 685k  & 4,100k      & 3,915k & 851k 
                                                       \\
Time [ms]     & 22.3     & 4.8    & 169.1   & 41.3       & 7.5     & 12.4  
                                                      \\

\hline
\end{tabular}
\end{adjustbox}
\end{center}
\label{table:speedparam}
\end{table}

\subsection{User Modification Comparisons}
\label{sec:comp_dni}
We demonstrate the effects of the modification of the texture gating strength on the denoising results. We use the GTCNN-D1 model and shift $\e_2$ by $\lambda$. The values of the modification parameter $\lambda$ ranged from -0.5 to 0.5 at an interval of 0.25 for the visual comparison. We also compare the modification results with the previous feature modification method called AdaFM-Net \cite{AdaFM}.  AdaFM-Net allows us to control the denoising strength by a parameter $\alpha$. The main component of AdaFM-Net is based on DnCNN \cite{DNCNN} and adopts a feature modification layer after each convolution layer.  To achieve configurable image denoising, AdaFM-Net takes two steps training strategy. In the first step, the entire network is trained with a noise level of $\sigma=15$. Then, as the second step, the feature modification layers are trained for a  noise level of $\sigma=75$. In our experiment, the interpolation parameter $\alpha$ ranged from 0 to 1 at an interval of 0.25. Note that our GTCNN modifies the texture strength unlike AdaFM-Net, which modifies the denoising strength.

%% figure 1,comp: compare with DNI

\begin{figure}[t]
  \centering
  \includegraphics[width=\linewidth]{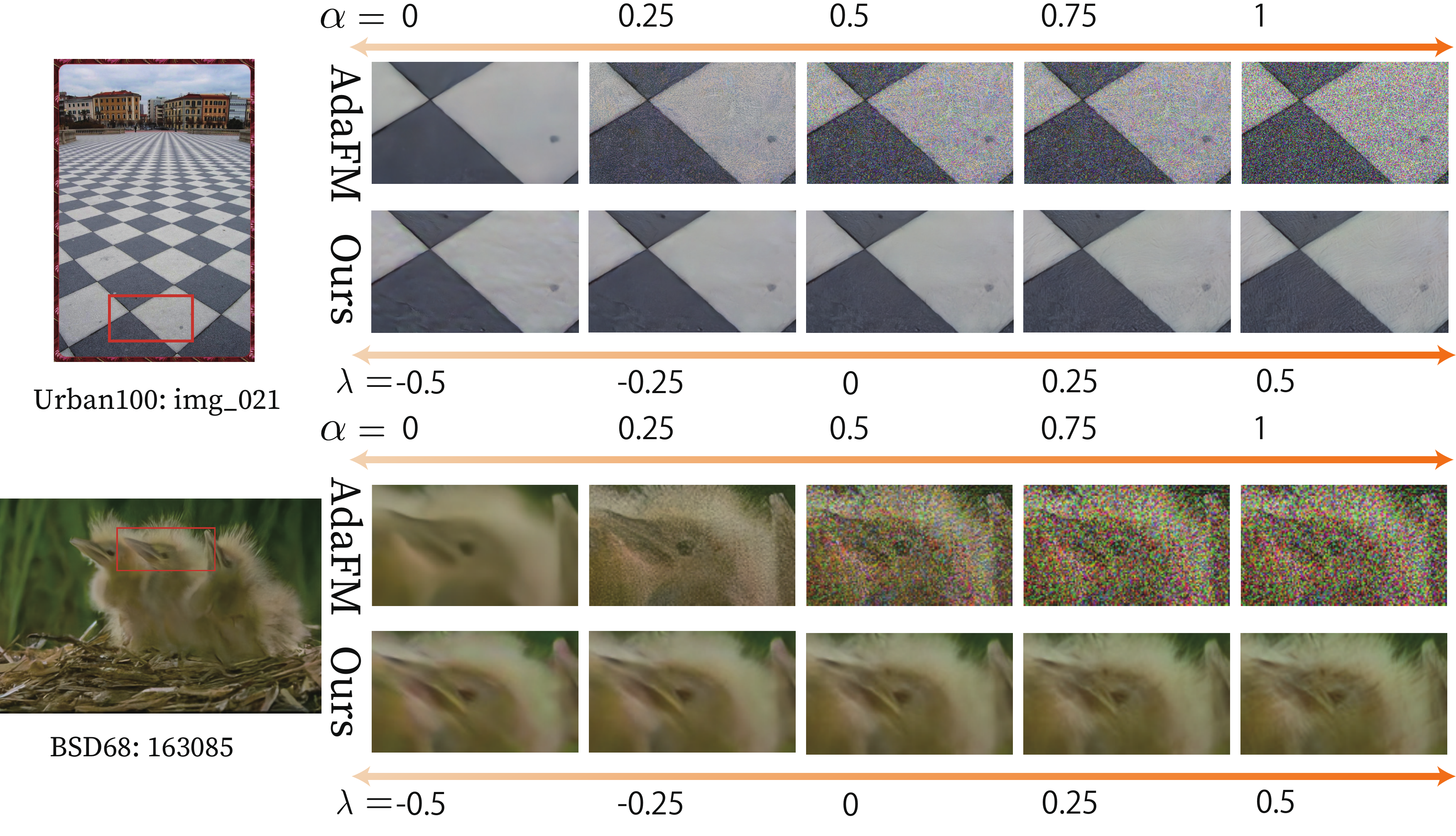}

      \caption{Comparison of the performance of our gate modulation method and that of AdaFM-Net \cite{AdaFM} for a noise level of $\sigma$ = 70
}
  
  \label{fig:mod_comp_dni}
\end{figure}

\figref{fig:mod_comp_dni} shows a few examples of the GTCNN-D1 and AdaFM-Net results. The resulting images show that AdaFM-Net can control the denoising strength, but it also produces some artifacts when the denoising strength is weak. Our GTCNN-D1 model demonstrates a gradual change in the texture strength of the output image with an increase in $\lambda$. These results show that modification of our GTCNN-D1 model is more flexible and yields fewer artifacts than AdaFM-Net. Notably, the GTCNN model does not require any additional training, modules, or networks.

% -----------------------------------------------------------------------
% ------------------------------Ablation studies-----------------------------------------
% -----------------------------------------------------------------------
\subsection{Ablation Studies}
\label{sec:ablation}
In this subsection, we show the effectiveness of our design choice. We first evaluate our gating mechanism and then extensively evaluate the effect of a large receptive field. Furthermore, we perform ablation studies for texture strength modulation. All the denoising results are on BSD68\cite{BSD68} with a noise level of $\sigma$ = 50.
Note that for ablation studies, we use a part of the small training set, which consists of 1,600 patches.

\subsubsection{Comparison with the Other Modules.}
\label{sec:com_modules}

To achieve a better trade-off between the denoising performance and computational resources, we compared our
GTCNN-D1 model with several attention modules and demonstrate their performance based on image recognition \cite{SENet,CBAM} and image restoration \cite{RCAN,RNAN,RidNet}. These attention modules are the squeeze and excitation (SE) module \cite{SENet}, the convolutional block attention module (CBAM) \cite{CBAM}, the NL module \cite{NL}, and the residual local attention block (RLAB) \cite{RNAN}. We incorporate these modules into each intermediate layer of a DnCNN \cite{DNCNN}, which is the baseline denoising method.
Note that the NL module \cite{NL} is assigned only to one intermediate layer of the baseline to suppress memory consumption; the NL module consumes an enormous amount memory.
We also provide the results of the GTCNN-D1 model with a sigmoid gate.

\tabref{table:Ablation_Attention} clearly shows that only our softmax gate can significantly improve the denoising accuracy with a short runtime and small number of parameters.

\subsubsection{Effectiveness of a Large Receptive Field.}
\label{sec:rec_abl}
To show the effectiveness in the global context, we evaluate the denoising results by the GTCNN-D1 model with a different number of stages for the GTL.
We denote a GTL with $S$ stages by GTL-B$S$. For example, GTL-B0 means the GTL without downsampling operations ($S=0$). We evaluated the six GTLs with different values of S: GTL-B0, GTL-B1, GTL-B2, GTL-B3, GTL-B4, and GTL-B5.

Table 2 shows the performance of all the GTCNN models.
The results clearly show that the large receptive field significantly improves the denoising results of the GTCNN model. Notably, GTL-B1 improves the performance of GTL-0 by utilizing only two additional convolution layers with a small spatial resolution. We decided to use GTL-B4 as the default setting of the GTL from this result.

\setlength{\tabcolsep}{3mm} 
%% Tabel 1, Ablation Study 1 : compare with attention
\begin{table}[!t]
\caption{\textbf{Comparison of DnCNN models with widely used attention mechanisims}. We observe that our simple gating mechanism with a softmax gate surprisingly outperforms recently proposed attention mechanisms \cite{SENet,CBAM,NL,RNAN} } 

\begin{center}
\begin{adjustbox}{max width=\textwidth}
\begin{tabular}{ l|c|c|c }
\hline
Description & PSNR [dB] & Number of Parameters & Time [ms]  \\
\hline
\hline
DnCNN (baseline) \cite{DNCNN}    & 25.90       & 555k
                                                    & 22.3   \\
DnCNN + SE \cite{SENet}   & 25.86        & 564k          
                                                    & 28.0   \\
DnCNN + CBAM \cite{CBAM}   & 26.09        & 566k          
                                                    & 670.8   \\
DnCNN + NL \cite{NL}   & 26.10        & 566k          
                                                    & 14,738.7   \\
DnCNN + RLAB \cite{RNAN}   & 26.21        & 5604k          
                                                    & 162.1   \\
\hline
GTCNN-D1 $\delta= \rm sigmoid$                 & 26.23   
                                                    &  851k  &   12.4 \\
GTCNN-D1  $\delta=\rm softmax$                 & \textbf{26.30}   
                                                    &  851k  &   12.4\\
\hline
\end{tabular}
\end{adjustbox}
\end{center}

\label{table:Ablation_Attention}
\end{table}

%% Tabel 3, Ablation Study 3 : Effectiveness of large receptive filed
\begin{table}[t]
\caption{\textbf{Effect of the number of stages in the GTL on the performance}. We observe that incorporating the global context of the scene significantly boosts the denoising accuracy} 

\begin{center}
\begin{adjustbox}{max width=\textwidth}
\begin{tabular}{ l|c|c|c }
\hline
Description & PSNR [dB] & Number of Parameters & Time [ms]  \\
\hline
\hline
GTCNN-D1 with GTL-B0       & 25.27       & 186k
                                                    & 8.9   \\
GTCNN-D1 with GTL-B1        & 26.03        & 297k         
                                                    & 10.9  \\
GTCNN-D1 with GTL-B2        & 26.24        & 481k        
                                                    & 11.9   \\
GTCNN-D1 with GTL-B3        & 26.27        & 666k         
                                                    & 12.2   \\
GTCNN-D1 with GTL-B4        & 26.30        & 851k         
                                                    & 12.4   \\
GTCNN-D1 with GTL-B5        & 26.31        & 1,003k         
                                                    & 12.7  \\

\hline
\end{tabular}
\end{adjustbox}
\end{center}

\label{table:Ablation_stage}
\end{table}
\subsubsection{Analysis of Texture Modulation on the GTL.}
\label{sec:mod_abl}
\begin{figure}[t]
  \centering
  \includegraphics[width=\linewidth]{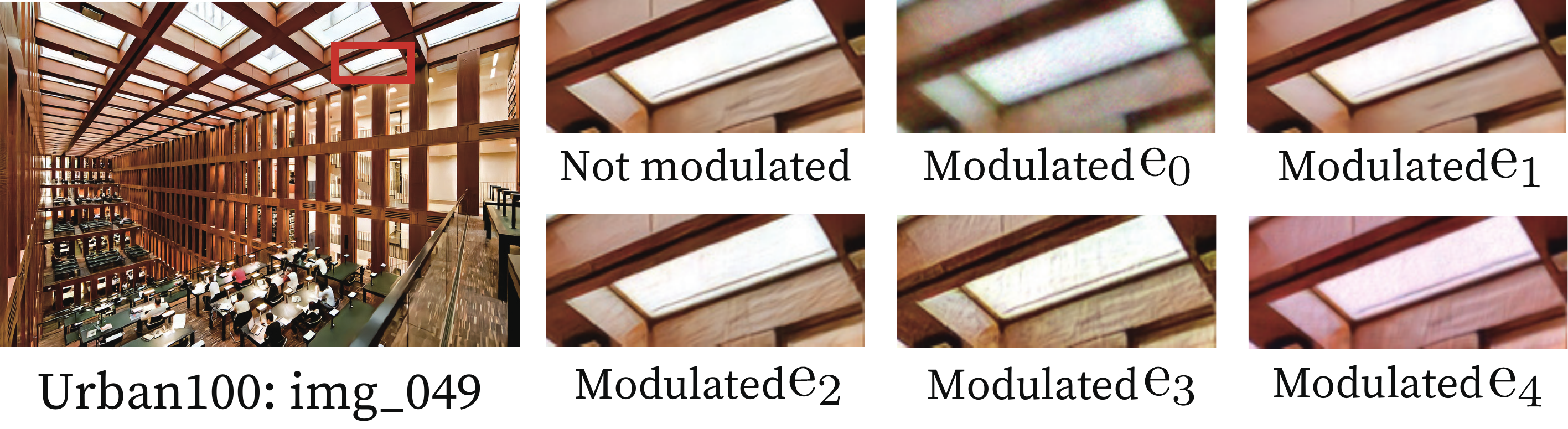}

\caption{\textbf{Results of the modulation by shifting different skip connections.} The results show that the GTCNN model can modulate the texture gating strength without artifacts. All the $\e_s$ values were shifted by $\lambda=0.5$}

  \label{fig:abl_mod}
 
\end{figure}

To control the texture gating strength, we shift the output of the skip connection by $\lambda$, as shown in \figref{fig:GTL}. We modulate each $ \e_0$, $\e_1$, $ \dots$, $\e_5$ one at a time.
The qualitative comparisons are shown in \figref{fig:abl_mod}. The results show that our GTCNN-D1 model can control the texture strength in the denoising results without producing artifacts.
We can observe that the texture of the denoising results is modulated naturally. Thus, we choose $\e_2$ as the default skip connection to be modulated.

\section{Conclusions}

Previously proposed denoising methods tend to output smooth results, which are caused by removing the texture information from the input image. To overcome this problem, the previous state-of-the-art methods employ an enormous number of parameters, which leads to high computational costs.
In contrast, we proposed a simple yet highly effective gated texture convolutional neural network (GTCNN), which removes texture information from the intermediate feature map. The GTCNN model achieved state-of-the-art performance with an order of magnitude fewer parameters than the previous state-of-the-art methods.
Furthermore, the GTCNN model allows us to interactively control the texture strength in the output image without any additional modules, training, or computational cost.

\clearpage
\bibliographystyle{splncs04}
\bibliography{Gated_Texture_CNN}
\end{document}